\documentclass{article}
\usepackage{spconf,amsmath,graphicx}
\usepackage{color}
\usepackage{amssymb}
\usepackage{booktabs}

\title{Robust copy-move forgery detection by false alarms control}
\name{Thibaud Ehret \thanks{Work supported by IDEX  Paris-Saclay IDI 2016, ANR-11-IDEX-0003-02, ONR  grant N00014-17-1-2552,  CNES MISS project, DGA Astrid ANR-17-ASTR-0013-01, DGA Defals challenge ANR-16-DEFA-0004-01, and MENRT.}}
\address{CMLA, ENS Cachan, CNRS, Universit\'e Paris-Saclay, 94235 Cachan, France}
\begin{document}
\maketitle
\begin{abstract}
 Detecting reliably copy-move forgeries is difficult  because  images do contain similar objects. The question is : how to discard natural image self-similarities while still detecting copy-moved parts as being ``unnaturally similar''? Copy-move may have been performed after a rotation, a change of scale and followed by JPEG compression or the  addition of noise. For  this reason, we base our method on SIFT,  which provides sparse keypoints with scale, rotation and illumination invariant descriptors. To  discriminate natural descriptor matches from artificial ones, we introduce an \textit{a contrario} method which gives theoretical guarantees on the number of false alarms. We validate our  method on several databases. Being fully unsupervised it can be integrated into any generic automated image tampering detection pipeline. %
\end{abstract}
\vspace{-0.5em}
\section{Introduction}
\label{sec:intro}
 Photo and video editing includes the insertion or removal of parts of the image, often performed by internal or external copy-move operations. The Poisson editing technique \cite{perez2003poisson,ipol.2016.163} allows for seamless insertions and 
 is now routinely used for special effects in movies, in software like Photoshop or in popular mobile phone applications.  Most editing operations are  driven by aesthetic goals. Yet their usage can easily become malicious and help forging false evidence, fake news, or alter results in scientific publications \cite{bik2016prevalence}. %

It is therefore of primary importance to provide public and professionals with reliable scientific tools detecting traces  of any intentional alteration of a photograph. Several different techniques are relevant here: image splicing (internal or external) can be detected through its local alterations of the compression encoding and of the JPEG blocks \cite{lin2009fast,cao2012robust,nikoukhah2018automatic}, its inconsistent  demosaicking traces \cite{popescu2005exposing,ferrara2012image,bammey2018automatic}, or directly  \cite{ng2004blind,hsu2007image,huh2018fighting}. Methods tracking other features such as noise inconsistencies,  lightning inconsistencies, chromatic aberration inconsistencies, etc. were listed in the broad review  \cite{farid2009image}. 

Our paper focuses on a specific type of image splicing called ``copy-move". As its name indicates it consists in copying a region of the image and pasting it somewhere else. Rotation, scaling, change of contrasts and other manipulations are sometimes applied to the piece being copied before pasting it. The method can be used to replicate objects, but sometimes also to hide an object by a texture borrowed elsewhere in the image.
Copy-move detection methods can be divided into two main categories: Block-based and keypoint-based. The block-based approaches try to match regions by blocks. 
In order to match the blocks more easily and more efficiently it is frequent to represent the block in  a compact form by dimensionality reduction, e.g. with PCA,  DCT \cite{cao2012robust} or DWT \cite{li2007sorted}. The  compact representation may also ensure the invariance of the detection to rotations by using Zernike moments \cite{cozzolino2015efficient,ipol.2018.213} or a similarity invariance with the Fourier-Mellin transform \cite{bayram2009efficient,li2010rotation}. These methods generally manage to detect  the forged regions, but are computationally demanding.
Instead of directly trying to match blocks, featured-based methods compute sets of keypoints and then match these keypoints. Many of these methods are based on SIFT \cite{amerini2010geometric,ardizzone2010detecting} or SURF \cite{bo2010image}. The descriptors associated to the keypoints are  invariant to rotation, scaling and even moderate affine distortions. Yet, precisely out of too much robustness, these methods may cause false detections when similar objects are present in an image. As argued in \cite{wen2016coverage}, most methods therefore suffer from a false positive  problem caused by the occurrence of ``natural'' self-similarity. This is the problem that we attempt to tackle here.

Section \ref{sec:method}  introduces our method and Section \ref{sec:experiments}  shows experimental results on different datasets. Finally perspectives are going to be presented in Section \ref{sec:conclusion}.

\vspace{-0.5em}

\section{Copy-move matching with SIFT-like matching}
\label{sec:method}

Like in the SIFT algorithm   \cite{lowe1999object} we start by computing a set of sparse keypoints. These keypoints are usually located in textured regions. Then a descriptor is associated to each of these keypoints. Finally the descriptors are matched to each other to define the detection. These three steps are summarized in the next paragraphs.

\vspace{-0.5em}

\subsection{Keypoints}

The keypoints correspond to the extrema of the normalized Laplacian scale-space. In practice they are computed using differences of Gaussians. The positions of the maxima are then found for each scale. To each keypoint is associated a scale and a principal orientation. A more detailed analysis can be found in \cite{ipol.2014.82}.

\vspace{-0.5em}

\subsection{Descriptors}

This first, classic, SIFT step gives a list $\mathcal{K} = (k_i)$ of keypoints. From each of these keypoints (consisting of a spatial position, a scale and an orientation), a square patch $p_i$ of size $(N+2) \times (N+2)$ can be sampled. The gradients in both directions are then computed from these patches yielding an $N \times N$ gradient patch $\mathcal{D}^i$ with vector values  $(\frac{\partial p_i}{\partial x}, \frac{\partial p_i}{\partial y})$. 

Contrary to SIFT, we keep these matrices for the matching step.  Indeed, using  histograms of gradients (HOGs) to represent the gradient patch would be too robust a representation and lead to the detection of natural repetitions. Hence, following \cite{von2015contrario} and \cite{rodriguez2018affine}. we encode  the key point $k_i$ by its gradient patch $\mathcal D_i$. This allows for an invariance to uniform illumination changes. In SIFT, the descriptors are computed on a grayscale version of the image for matching applications. In the forgery case, it is interesting to consider all information available.  So our gradient descriptors keep three channels, one for each color. For simplicity our  matching step will be presented using a grayscale descriptor, but extends immediately to color as well. Color descriptors will be used for the experiments in Section \ref{sec:experiments}. 

\vspace{-0.5em}

\subsection{Matching}
\label{sec:matching}

Two naturally similar objects are rarely  exactly similar. This is because there are always differences in their illumination in a real scene, and physical differences that do not necessarily catch the eye, in addition  to the acquisition  noise. Our matching process  takes advantage of these serious variations to discriminate between similar objects and digital copies.  Consider two keypoints $k_i$ and $k_j$ located respectively on the original object and on the forged copy, so that they match with a regular matching method such as SIFT  \cite{lowe1999object} or \cite{rodriguez2018affine}. In that case the descriptors $\mathcal D^i$ and $\mathcal D^j$ associated to these keypoints should be \emph{exactly} the same, namely $\forall k,l \in \{1,\dots,N\}, \mathcal{D}^i_{k,l} = \mathcal{D}^j_{k,l}$.
Of course this perfect quality  is not reached in practice.  Several copy-move steps  could introduce small differences such as: the interpolation due to a rotation or zoom or even a post-processing step such as the addition of noise and/or a compression after  forgery. Nevertheless, we can  enforce a very close match between each part of the descriptors by an exigence  like $\forall k,l \in \{1,\dots,N\}, \|\mathcal{D}^i_{k,l} - \mathcal{D}^j_{k,l}\|^2_2 \leqslant \tau$.
For the distance $d_{max}$ defined by
\begin{equation}
    d_{max}(\mathcal{D}^i, \mathcal{D}^j) = \max_{k,l \in \{1,\dots,N\}} \|\mathcal{D}^i_{k,l} - \mathcal{D}^j_{k,l}\|^2_2,
\end{equation}
the  suspicious match  test is simply $d_{max}(\mathcal{D}^i, \mathcal{D}^j) \leqslant \tau$.

\begin{figure}
    \centering
    \includegraphics[width=0.49\linewidth]{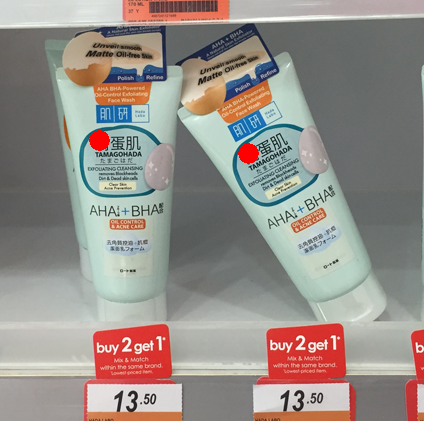}
    \includegraphics[width=0.49\linewidth]{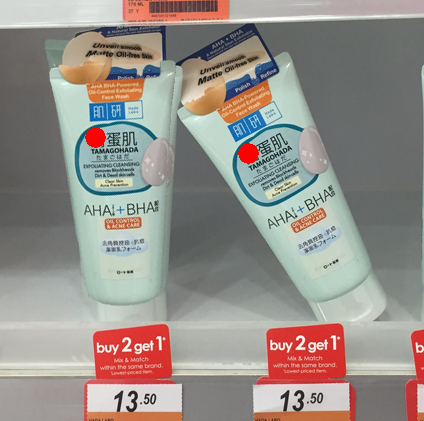}
    \includegraphics[width=0.24\linewidth]{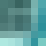}
    \includegraphics[width=0.24\linewidth]{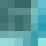}
    \includegraphics[width=0.24\linewidth]{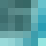}
    \includegraphics[width=0.24\linewidth]{images/distinguish/patch_orig.png}
    \caption{On the left image the two objects are similar but not digital copies of each other. On the right, one is a digital copy of the other. The patches shown below each respective image correspond to the red dots in the images. They show that a difference is visible at this level and therefore the descriptors can be discriminated. This is why the detection method can discard  genuinely similar objects.}
    \label{fig:distinguish}
\end{figure}
The key question is to fix  the  right detection threshold $\tau$, to have a matching criterion that rejects genuinely similar objects  while still detecting well copy-move forgeries. This threshold  can be computed rigorously   using the \emph{a-contrario} theory~\cite{desolneux2007gestalt} which is a probabilistic formalization of the \emph{non-accidentalness} principle~\cite{lowe1985perceptual}. This principle has shown its practical use for detection purposes  such as segment detection~\cite{grompone2010lsd}, vanishing points detection~\cite{lezama2014finding} and  anomaly detection~\cite{davy2018reducing}. The \emph{a-contrario} theory provides a way to compute automatically detection thresholds while having a control on the number of false alarms (NFA). It replaces the usual $p$-value by drawing into account the number of tests and therefore controlling the overall number of false alarms in a given detection  task.
The method only requires a simple \textit{a contrario} stochastic model on the perturbation. We will consider for now that $\mathcal{D}^i$ and $\mathcal{D}^j$ are derived from the same patch but one of them has been corrupted by Gaussian noise of variance $\sigma^2$ \textit{i.e.} since the descriptors consist of gradients $\forall k,l, \mathcal{D}^j_{k,l} = \mathcal{D}^i_{k,l} + n_{k,l}$ where $n_{k,l} \sim \mathcal{N}(0,2\sigma^2)$ and are independent. Matching both descriptors requires that $\max_{k,l} n_{k,l}^2 \leqslant \tau$.
The probability of matching in this case is then 
\begin{equation}
    \mathbb{P}\left(\max_{k,l} n_{k,l}^2 \leqslant \tau\right) = \prod_{k,l} \mathbb{P}\left(n_{k,l}^2 \leqslant \tau\right)
    = \mathbb{P}\left(n \leqslant \frac{\tau}{2\sigma^2}\right)^{N^2}
\end{equation}
where $n$ follows a $\chi^2$ distribution with $1$ degree of freedom. We can therefore control the number of false detections by choosing the proper $\tau$ according to 
\begin{equation}
    \tau = 2\sigma^2 \times chi2inv\left(\sqrt[N^2]{\frac{\epsilon}{N_{tests}}}\right),
    \label{eq:threshold}
\end{equation}
where $\epsilon$ is the number of false alarms per $N_{tests}$ number of tests and $chi2inv$ is inverse of the $\chi^2$ cumulative distribution function.
The main point of formula \eqref{eq:threshold} is that it reduces the initial method dependency on many detection parameters to just one, namely $\sigma$.  We can argue that this last one is not critical. Indeed, even though the dependency on $\sigma$ is strong, as long as the degradation is not too large, there will be a scale in which $\sigma$ is small enough so the detection will work: indeed $\sigma$ is  divided by two at each octave in the SIFT  method. For example, this exigent threshold can work for a noise of  4, but requires the tampered area to be four times larger for a detection. It  might  be objected that zooming down also makes naturally similar objects become more  similar. Yet our experiments indicate that this is not the case, indeed  their small but significant differences encompass all scales.
To summarize, granting that we allow for one false detection on average on a set  of images, the method is parameterless as it adapts to the number  of  tests and to the patch size. Of course it  might be coupled with an automatic noise estimator to give an good  guess of $\sigma$.
Assuming a perturbation noise of variance $\sigma^2=1$ and the  use of descriptors  of size $4 \times 4 \times 3$ (derived from $6\times6$ color patches), a number of false alarms of $\epsilon=1$ and testing on $100$ images with on average $50$ keypoints (this corresponds to the COVERAGE dataset presented in Section \ref{sec:experiments}), Equation \eqref{eq:threshold} gives $\tau=2.9$. The advantage of using color descriptors in this case is either to increase the size of the descriptor (allowing for a larger $\tau$ and detecting more) or to reduce the spatial size for a same size of descriptor (allowing to detect smaller forgeries). 

An interesting side effect is that this test is really fast to compute. Indeed to detect forgeries each keypoint must to  compared against all others. Since we are comparing keypoints inside a single image this gives $\frac{(K-1)(K-2)}{2}$ pairs to be tested, where $K$ is the number of descriptors.
(Of course all descriptor self-matches are discarded).   For large images the computation of the distance becomes quickly a bottleneck for distances that are costly. In our case it is not necessary to compute $d_{max}$ before doing the test, the test can be done during the computation of $d_{max}$ which allows for early stopping. Since most keypoints won't match, the number of operations done per comparison is in practice much smaller than the size $N^2$ of the descriptor. An experimental verification of this fact  is made in Section \ref{sec:experiments}.

Finally we need to take into account all possible flips for the forged regions. While the matching process doesn't detect flips it is possible to still detect them at the cost of a few more computations. The modified distance to test flips is then 
\begin{equation}
    d_{flip}(\mathcal{D}^i, \mathcal{D}^j) = \max_{k,l} \left(x^i_{k,l} - x^j_{k,l}\right)^2+ \left(y^i_{k,l} + y^j_{N-k+1,l}\right)^2 
\end{equation}
where $D_{k,l} = (x_{k,l},y_{k,l})$.
Indeed when flipped, the indexes in one direction are reversed but also the gradients in that direction are opposite. Thanks to the rotation invariance all flips are taken into account by just testing the flip in one direction (in our case in the $y$ direction). In the end, we test each pair of keypoints with both distances to take into account flips. Having to do twice the computation is not a problem in practice as each test is very efficient.

\section{Experiments}
\label{sec:experiments}

\begin{table}
    \caption{Detection statistics on the different datasets compared to the reported results from \cite{wen2016coverage}. The proposed method achieves similar true positive detections for a very limited number of false detections. The only false detection is shown in \ref{fig:failure}. The false detections are computed on the original images (with no forgeries).}
	\label{tab:detection}
	\begin{center}
		{%
        \begin{tabular}{ @{}l  l @{\hskip 1cm} c  c@{}}
            \toprule
            Dataset     & Method   & \shortstack{True\\ detections}      &  \shortstack{False\\ detections}   \\
            \midrule  
            COVERAGE    & Proposed &  43\%   &  1\%  \\
                        & Previous &  50.5\% &  -     \\
            \midrule  
            GRIP        & Proposed &  70\% &  1.3\%  \\
                        & Previous &  71\%   &  -     \\
            \midrule  
            IM         & Proposed &  81.2\% &  0\%  \\
                        & Previous &  75\%   &  -     \\
            \midrule 
            IM JPEG80  & Proposed &  64.6\% &  0\%  \\
            \midrule
            IM NOISE20 & Proposed &  79.2\% &  0\%  \\
            \bottomrule
        \end{tabular}}
    \end{center}
\end{table}
In this section the images are all shown in grayscale even though they are originally in color. This allows for a better visualization of the matches. Nevertheless, the descriptors  were color descriptors. We present results on three different datasets: GRIP \cite{cozzolino2015efficient}, Image Manipulation (IM) \cite{christlein2012evaluation} and COVERAGE \cite{wen2016coverage} which is the dataset that inspired this study as it focuses on distinguishing forgeries from similar but genuine objects. All images shown in this section come from these datasets.

We decided to use descriptors of size $3\times8\times8$ for the IM dataset and $3\times4\times4$ otherwise as the images from the IM dataset are much larger than the ones from the other datasets. Indeed the size of the descriptors needs be chosen so to be smaller than the expected size of the forged regions. Each time the threshold was computed using Equation \eqref{eq:threshold} from Section \ref{sec:matching}. We also verified that the number of comparisons done to compare two descriptors was much smaller than the size of a descriptor. For example for the image shown in Figure \ref{fig:distinguish}, of size $424 \times 421$ and containing $207$ descriptors only $1.1$ comparisons were necessary on average for a descriptor size of $3\times6\times6=108$ that is almost the size of the descriptor used for SIFT. Thus the detection is really fast even for large images with a large number of keypoints.

\begin{figure}
    \centering
    \includegraphics[width=0.49\linewidth]{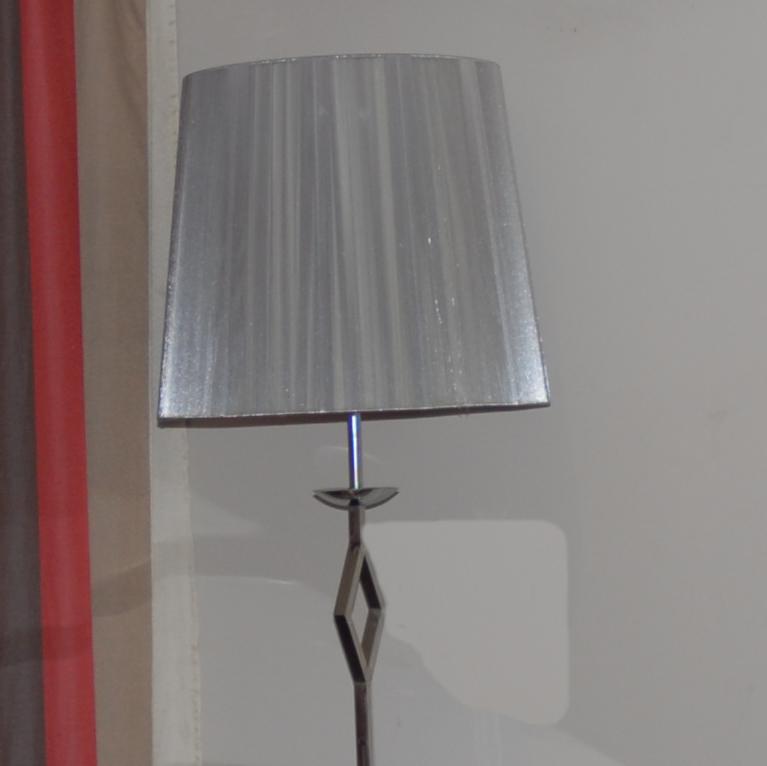}
    \includegraphics[width=0.49\linewidth]{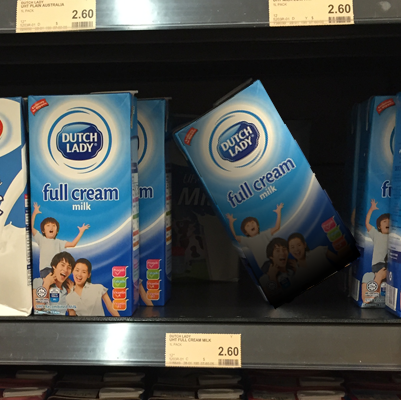}
    \includegraphics[width=0.49\linewidth]{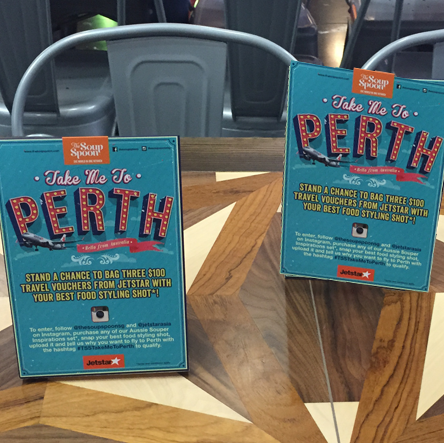}
    \includegraphics[width=0.49\linewidth]{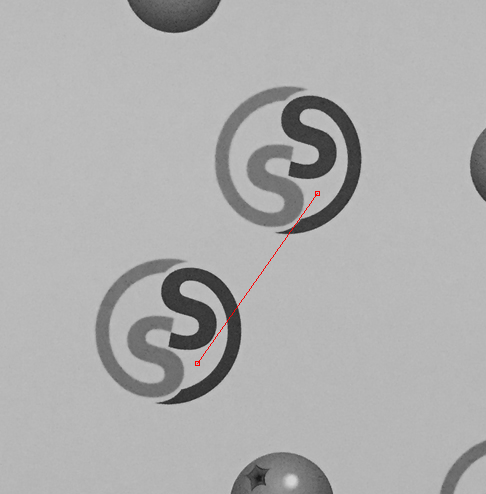}
    \caption{Examples of forgeries that were not successfully detect in the different datasets. From top to bottom, left to right: an example where the forged region is completely flat, with a non-uniform change of illumination, with a tilt applied and the false detection.}
    \label{fig:failure}
\end{figure}

Table \ref{tab:detection} shows that while the methods focuses on being robust to similar objects and reduces as much as possible false detections, it is actually competitive with previous keypoint  based methods. Moreover, the number of false alarm is definitely under control : only very little false alarms were found in all three datasets. One of these false detection is shown in Figure \ref{fig:failure}. We also verified that while robust to similar objects (and therefore very precise) the method  still is robust to reasonable noise and compression. 

\begin{figure}
    \centering
    \includegraphics[width=0.49\linewidth]{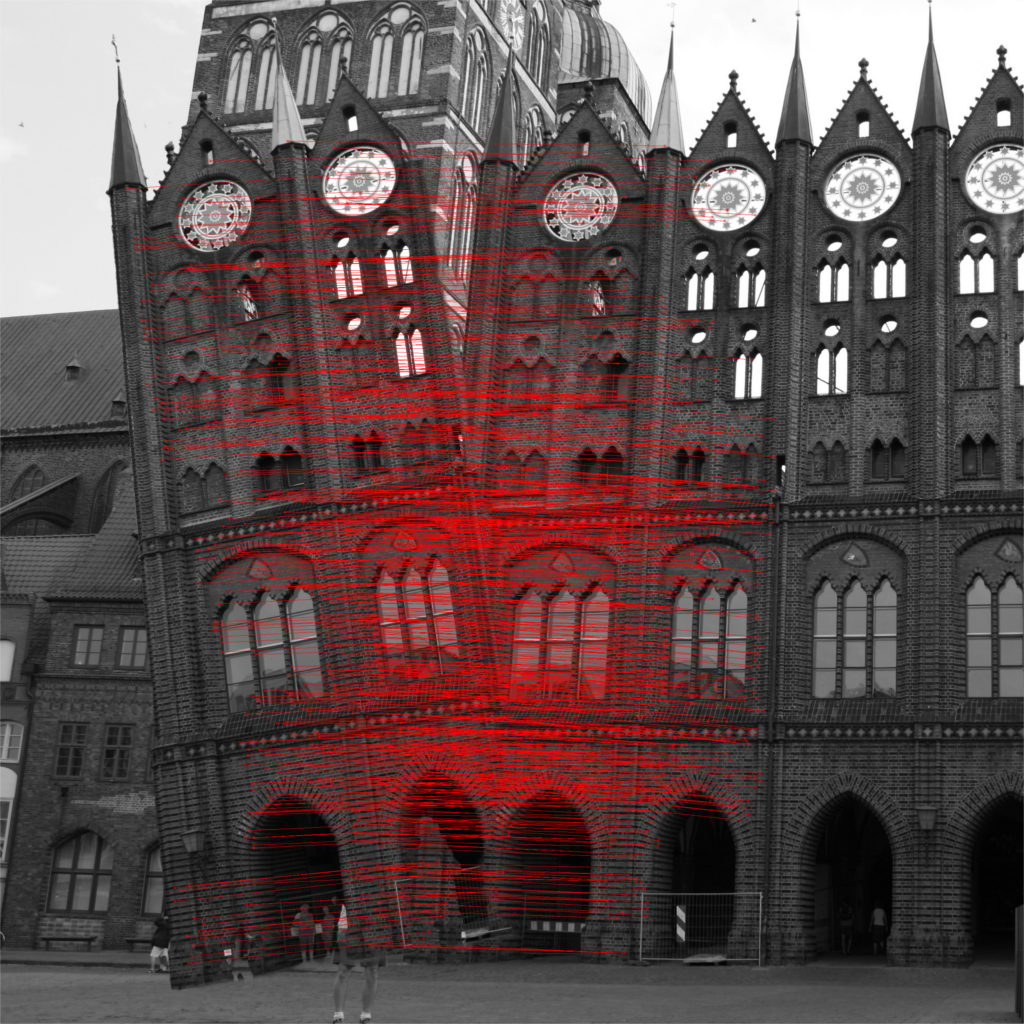}
    \includegraphics[width=0.49\linewidth]{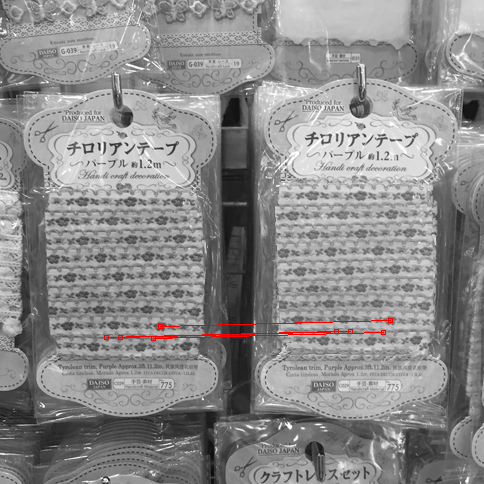}
    \includegraphics[width=0.49\linewidth]{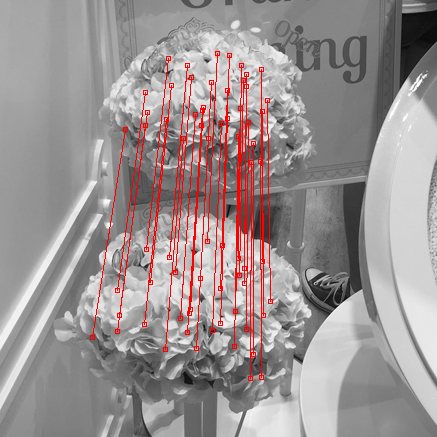}
    \includegraphics[width=0.49\linewidth]{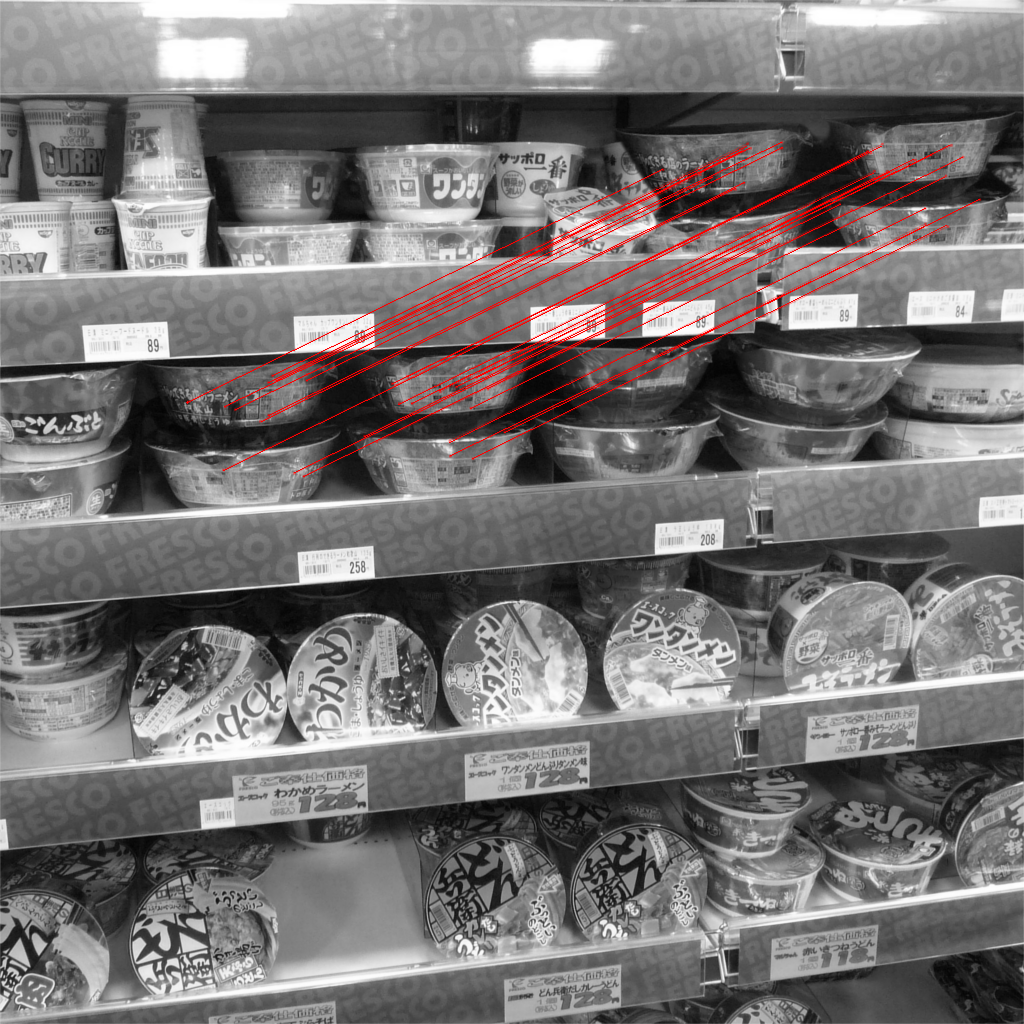}
    \caption{Examples of forgeries that were successfully detected in the different datasets. From the top to bottom, left to right: an example with a rotation, a change of illumination, a change of scale and with JPEG compression.}
    \label{fig:success}
\end{figure}

Figure \ref{fig:success} shows different examples of successful detections. The method is able to detect well rotation, uniform illumination changes, scaling and compression. As can be seen, the more texture the forged region has the easier it is to detect. This is because a textured region will generate more keypoints and therefore will increase its chances of matching. 

Figure \ref{fig:failure} shows several failure examples. Most failures come from the fact that the method can't deal with more severe distortions such as a tilt or non-uniform illumination change. The method also fails to detect flat regions, as no keypoints are computed on these regions. As for the false detection, it does not contradict the  a contrario model. We requested at most $\epsilon=1$ false detection per $100$ images with the threshold given in Section \ref{sec:matching}, and we  found one with $200$ images tested for the COVERAGE dataset.

\section{Perspectives}
\label{sec:conclusion}

In this paper we have presented an unsupervised method to detect copy-move forgeries that is not only invariant to rotation, scaling and global change of illumination, but also  robust to the presence of similar but genuinely different objects or regions. The method,  being parameter-less  and very fast,  can be included in the necessary long series of tampering tests applied to a suspicious image.   

The limits of the method are closely linked to its strength. Because it is  robust to the presence of naturally similar objects, it is less reliable in case of large degradation of a copied digital ones.  We nevertheless found that  the method  is robust enough to usual noise and compression levels. An image that has been degraded too much is suspicious anyway, since nowadays the quality of an image taken with a mobile is very good.  Thus only images with a good enough quality should be tested.  Highly degraded images would be  anyway suspicious regardless of  any such more  sophisticated examination. The second limit is the usage of sparse keypoints. These keypoints are only computed in regions that are contrasted enough (non-flat areas) which means that forgeries in these regions might not detected. Finally matching keypoints give anchor points and do not delimit forged regions precisely.  A natural extension  of the  method  would be to extract the forged regions from the anchor points while still keeping a good control over the number of false detections. Finally coupling the method with a noise  estimator could arguably make it still more discriminant.

\ninept
\bibliographystyle{IEEEbib}
\bibliography{main}

\begin{thebibliography}{10}

\bibitem{perez2003poisson}
P.~P{\'e}rez, M.~Gangnet, and A.~Blake,
\newblock ``Poisson image editing,''
\newblock {\em ACM TOG}, 2003.

\bibitem{ipol.2016.163}
J.~M. Di~Martino, G.~Facciolo, and E.~Meinhardt-Llopis,
\newblock ``{Poisson Image Editing},''
\newblock {\em {IPOL}}, vol. 6, pp. 300--325, 2016.

\bibitem{bik2016prevalence}
E.~Bik, A.~Casadevall, and F.~Fang,
\newblock ``The prevalence of inappropriate image duplication in biomedical
  research publications,''
\newblock {\em MBio}, vol. 7, no. 3, pp. e00809--16, 2016.

\bibitem{lin2009fast}
Z.~Lin, J.~He, X.~Tang, and C.-K. Tang,
\newblock ``Fast, automatic and fine-grained tampered jpeg image detection via
  dct coefficient analysis,''
\newblock {\em Pattern Recognition}, vol. 42, no. 11, pp. 2492--2501, 2009.

\bibitem{cao2012robust}
Y.~Cao, T.~Gao, L.~Fan, and Q.~Yang,
\newblock ``A robust detection algorithm for copy-move forgery in digital
  images,''
\newblock {\em Forensic science int.}, vol. 214, no. 1-3, pp. 33--43, 2012.

\bibitem{nikoukhah2018automatic}
T.~Nikoukhah, R.~Grompone~von Gioi, M.~Colom, and J.-M. Morel,
\newblock ``Automatic jpeg grid detection with controlled false alarms, and its
  image forensic applications,''
\newblock in {\em IEEE MIPR}. IEEE, 2018, pp. 378--383.

\bibitem{popescu2005exposing}
A.~Popescu and H.~Farid,
\newblock ``Exposing digital forgeries in color filter array interpolated
  images,''
\newblock {\em Trans. on Signal Processing}, vol. 53, no. 10, pp. 3948--3959,
  2005.

\bibitem{ferrara2012image}
P.~Ferrara, T.~Bianchi, A.~De~Rosa, and A.~Piva,
\newblock ``Image forgery localization via fine-grained analysis of cfa
  artifacts,''
\newblock {\em IEEE TIFS}, vol. 7, no. 5, pp. 1566--1577, 2012.

\bibitem{bammey2018automatic}
Q.~Bammey, R.~Grompone~von Gioi, and J.-M. Morel,
\newblock ``Automatic detection of demosaicing image artifacts and its use in
  tampering detection,''
\newblock in {\em IEEE MIPR}. IEEE, 2018, pp. 424--429.

\bibitem{ng2004blind}
T.-T. Ng, S.-F. Chang, and Q.~Sun,
\newblock ``Blind detection of photomontage using higher order statistics,''
\newblock in {\em ISCAS}. IEEE, 2004, vol.~5.

\bibitem{hsu2007image}
Y.-F. Hsu and S.-F. Chang,
\newblock ``Image splicing detection using camera response function consistency
  and automatic segmentation,''
\newblock in {\em Int. conf. on Multimedia and Expo}. IEEE, 2007, pp. 28--31.

\bibitem{huh2018fighting}
M.~Huh, A.~Liu, A.~Owens, and A.~Efros,
\newblock ``Fighting fake news: Image splice detection via learned
  self-consistency,''
\newblock {\em arXiv preprint arXiv:1805.04096}, 2018.

\bibitem{farid2009image}
H.~Farid,
\newblock ``Image forgery detection,''
\newblock {\em IEEE Signal processing magazine}, vol. 26, no. 2, pp. 16--25,
  2009.

\bibitem{li2007sorted}
G.~Li, Q.~Wu, D.~Tu, and S.~Sun,
\newblock ``A sorted neighborhood approach for detecting duplicated regions in
  image forgeries based on dwt and svd,''
\newblock in {\em Int. conf. on Multimedia and Expo}. IEEE, 2007, pp.
  1750--1753.

\bibitem{cozzolino2015efficient}
D.~Cozzolino, G.~Poggi, and L.~Verdoliva,
\newblock ``Efficient dense-field copy--move forgery detection,''
\newblock {\em TIFS}, vol. 10, no. 11, pp. 2284--2297, 2015.

\bibitem{ipol.2018.213}
T.~Ehret,
\newblock ``{Automatic Detection of Internal Copy-Move Forgeries in Images},''
\newblock {\em {IPOL}}, vol. 8, pp. 167--191, 2018.

\bibitem{bayram2009efficient}
S.~Bayram, H.~Sencar, and N.~Memon,
\newblock ``An efficient and robust method for detecting copy-move forgery,''
\newblock in {\em ICASSP}. IEEE, 2009, pp. 1053--1056.

\bibitem{li2010rotation}
W.~Li and N.~Yu,
\newblock ``Rotation robust detection of copy-move forgery.,''
\newblock in {\em ICIP}, 2010.

\bibitem{amerini2010geometric}
I.~Amerini, L.~Ballan, R.~Caldelli, A.~Del~Bimbo, and G.~Serra,
\newblock ``Geometric tampering estimation by means of a sift-based forensic
  analysis.,''
\newblock in {\em ICASSP}, 2010.

\bibitem{ardizzone2010detecting}
E.~Ardizzone, A.~Bruno, and G.~Mazzola,
\newblock ``Detecting multiple copies in tampered images,''
\newblock in {\em ICIP}. IEEE, 2010.

\bibitem{bo2010image}
X.~Bo, W.~Junwen, L.~Guangjie, and D.~Yuewei,
\newblock ``Image copy-move forgery detection based on surf,''
\newblock in {\em MINES}. IEEE, 2010, pp. 889--892.

\bibitem{wen2016coverage}
B.~Wen, Y.~Zhu, R.~Subramanian, T.-T. Ng, X.~Shen, and S.~Winkler,
\newblock ``Coverage—a novel database for copy-move forgery detection,''
\newblock in {\em ICIP}. IEEE, 2016.

\bibitem{lowe1999object}
D.~Lowe,
\newblock ``Object recognition from local scale-invariant features,''
\newblock in {\em CVPR}. IEEE, 1999, vol.~2, pp. 1150--1157.

\bibitem{ipol.2014.82}
I.~Rey~Otero and M.~Delbracio,
\newblock ``{Anatomy of the SIFT Method},''
\newblock {\em {IPOL}}, vol. 4, pp. 370--396, 2014.

\bibitem{von2015contrario}
Rafael Grompone and Viorica P{\u{a}}tr{\u{a}}ucean,
\newblock ``A contrario patch matching, with an application to keypoint matches
  validation,''
\newblock in {\em ICIP}. IEEE, 2015.

\bibitem{rodriguez2018affine}
M.~Rodr{\'\i}guez and R.~Grompone~von Gioi,
\newblock ``Affine invariant image comparison under repetitive structures,''
\newblock in {\em IEEE ICIP}. IEEE, 2018.

\bibitem{desolneux2007gestalt}
A.~Desolneux, L.~Moisan, and J.-M. Morel,
\newblock {\em From gestalt theory to image analysis: a probabilistic
  approach}, vol.~34,
\newblock Springer Science \& Business Media, 2007.

\bibitem{lowe1985perceptual}
D.~Lowe,
\newblock {\em Perceptual organization and visual recognition},
\newblock Kluwer Academic Publishers, 1985.

\bibitem{grompone2010lsd}
R.~Grompone Von~Gioi, J.~Jakubowicz, J.-M. Morel, and G.~Randall,
\newblock ``Lsd: A fast line segment detector with a false detection control,''
\newblock {\em IEEE PAMI}, vol. 32, no. 4, pp. 722--732, 2010.

\bibitem{lezama2014finding}
J.~Lezama, R.~Grompone, G.~Randall, and J.-M. Morel,
\newblock ``Finding vanishing points via point alignments in image primal and
  dual domains,''
\newblock in {\em CVPR}, 2014.

\bibitem{davy2018reducing}
A.~Davy, T.~Ehret, J.-M. Morel, and M.~Delbracio,
\newblock ``Reducing anomaly detection in images to detection in noise,''
\newblock in {\em ICIP}. IEEE, 2018.

\bibitem{christlein2012evaluation}
V.~Christlein, C.~Riess, J.~Jordan, C.~Riess, and E.~Angelopoulou,
\newblock ``An evaluation of popular copy-move forgery detection approaches,''
\newblock {\em arXiv preprint arXiv:1208.3665}, 2012.

\end{thebibliography}

\end{document}